%% file: main.tex
\documentclass{article}

    \PassOptionsToPackage{numbers, compress}{natbib}

\usepackage[preprint]{neurips_2023}
\usepackage{enumitem}

\usepackage[utf8]{inputenc} 
\usepackage[T1]{fontenc}    

\usepackage{url}            
\usepackage{booktabs}       
\usepackage{amsfonts}       
\usepackage{nicefrac}       
\usepackage{microtype}      
\usepackage{xcolor}         
\definecolor{cvprblue}{rgb}{0.21,0.49,0.74}
\usepackage[colorlinks=true,citecolor=cvprblue,]{hyperref}
 \usepackage{graphicx}
\usepackage{caption}
\usepackage{amsmath} 
\usepackage{multirow}
\usepackage{orcidlink}
\usepackage{multirow}
 \usepackage{subcaption}

 \usepackage{bbding}
 \usepackage{pifont}
 \newcommand{\increase}[1]{
    \textsuperscript{\textcolor{green}{#1$\uparrow$}}
}
\newcommand{\cmark}{\ding{51}}%
\newcommand{\xmark}{\ding{55}}%
\usepackage{xspace}
\def\ours{SimC3D\xspace}  
\newcommand{\eg}{e.g.\xspace}
\newcommand{\ie}{i.e.\xspace}

\title{SimC3D: A Simple Contrastive 3D Pretraining Framework Using RGB Images}
\author{%
    Jiahua Dong$^{1\dagger}$, Tong Wu$^{2}$, Rui Qian$^{2}$, Jiaqi Wang$^{1}$ \\
    $^1$Shanghai AI Laboratory \\
    $^2$The Chinese University of Hong Kong \\
    \texttt{cnjiahuadong@gmail.com, wutong16.thu@gmail.com,} \\
    \texttt{qr021@ie.cuhk.edu.hk,  wjqdev@gmail.com}
}

\begin{document}

\renewcommand{\thefootnote}{\fnsymbol{footnote}}
\footnotetext[2]{~This work was completed during an internship.}
\maketitle
\input{sec/0_abstract}

\input{sec/1_intro}

\input{sec/2_related}
\input{sec/3_method}

\input{sec/4_experiment}

\input{sec/5_conclusion}


\bibliographystyle{unsrtnat}
\bibliography{ref}
\clearpage

\input{sec/X_suppl}



\end{document}

%% file: sec/0_abstract.tex
\begin{figure}[h]

    \centering
    \includegraphics[width=1.0\linewidth]{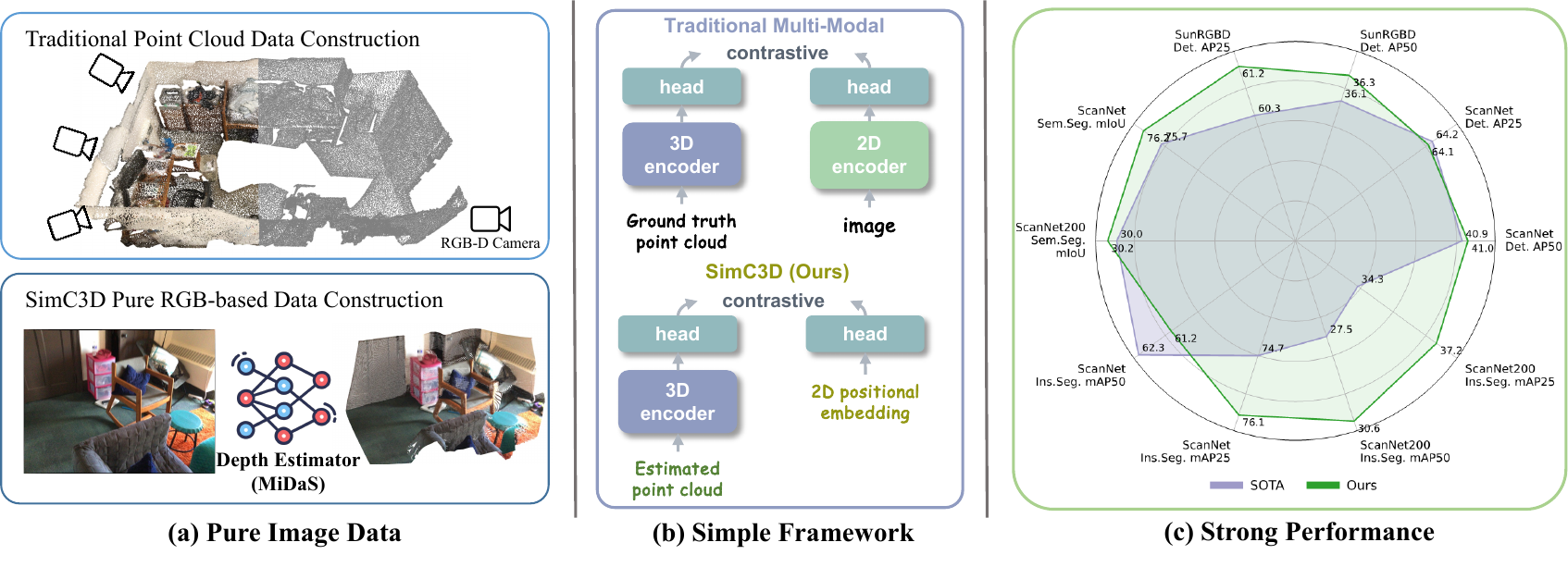}
    \vspace{-5mm}
    \captionof{figure}{\textbf{SimC3D performs contrastive 3D pretraining with three appealing properties.} \textbf{(1) Pure image data}: It simplifies the requirements of pre-training datasets, transitioning from expensive 3D point clouds to pure RGB images. \textbf{(2) Simple framework}: In contrast to previous multi-modal frameworks that rely on an additional 2D backbone, \eg, ResNet~\cite{resnet}, to encode image features,  SimC3D directly employs 2D positional embeddings as the training objective, thereby eliminating the need for a 2D encoder within the framework. \textbf{(3) Strong performance}: Although simplifying the data requirements and training framework, SimC3D can still achieve strong performance across various downstream tasks, as shown in the radar plot where `SOTA' represents the highest score achieved in prior works.}
    \label{fig:teaser}
\end{figure}
    \vspace{-2mm}

\begin{abstract}
The 3D contrastive learning paradigm has demonstrated remarkable performance in downstream tasks through pretraining on point cloud data. Recent advances involve additional 2D image priors associated with 3D point clouds for further improvement. Nonetheless, these existing frameworks are constrained by the restricted range of available point cloud datasets, primarily due to the high costs of obtaining point cloud data. To this end, we propose SimC3D, a simple but effective 3D contrastive learning framework, for the first time, pretraining 3D backbones from pure RGB image data. SimC3D performs contrastive 3D pretraining with three appealing properties. \textbf{(1) Pure image data:} SimC3D simplifies the dependency of costly 3D point clouds and pretrains 3D backbones using solely RBG images. By employing depth estimation and suitable data processing, the monocular synthesized point cloud shows great potential for 3D pretraining. \textbf{(2) Simple framework:} Traditional multi-modal frameworks facilitate 3D pretraining with 2D priors by utilizing an additional 2D backbone, thereby increasing computational expense. In this paper, we empirically demonstrate that the primary benefit of the 2D modality stems from the incorporation of locality information. Inspired by this insightful observation, SimC3D directly employs 2D positional embeddings as a stronger contrastive objective, eliminating the necessity for 2D backbones and leading to considerable performance improvements. \textbf{(3) Strong performance:} SimC3D outperforms previous approaches that leverage ground-truth point cloud data for pretraining in various downstream tasks. Furthermore, the performance of SimC3D can be further enhanced by combining multiple image datasets, showcasing its significant potential for scalability. The code will be available at \href{https://github.com/Dongjiahua/SimC3D}{https://github.com/Dongjiahua/SimC3D}.




\end{abstract}

%% file: sec/1_intro.tex
\section{Introduction}
Visual pre-training~\cite{simCLR,simsiam,mae} has rapidly developed in recent years, leading to significant improvements in downstream tasks such as image classification~\cite{resnet,imagenet} and object detection~\cite{fastrcnn}. These 2D pre-trained models can even surpass supervised models~\cite{simCLR,he2019moco,simsiam,mae}. Their notable strength is the ability to be directly applied to diverse images, enabling large-scale training~\cite{vit}. Inspired by this, 3D pre-training methods~\cite{pointcontrast,c_scene,depthcontrast,iae,invar3D,ponder,4dcontrast,strl,trans3d,mae3d,mae3dobj,maskd,randomroom,crosspoint} have emerged, showing promise in solving 3D downstream tasks. However, many self-supervised frameworks are limited by data requirements, especially point cloud data, hindering broader applicability.

Attempts have been made to reduce this dependency. Earlier contrastive-based works~\cite{pointcontrast,c_scene,maskd} relied on point cloud correspondences for contrastive learning, necessitating complete point cloud data. To address this, DepthContrast~\cite{depthcontrast} proposes a monocular point cloud pre-training method, treating each point cloud as a whole instance in contrastive learning. However, this requires a large batch size and an extended training period. More recently, multi-modal methods~\cite{crosspoint,invar3D} have exhibited promising performance with single-view data, establishing dense correspondence between point clouds and images but necessitating an additional 2D input and a robust 2D backbone.
Nevertheless, previous works still require a substantial amount of point cloud data. Although some just require single-view point cloud~\cite{depthcontrast, invar3D}, these data still need to be captured through depth sensors and known camera calibrations. This requirement makes data acquisition for diverse scenarios challenging and limits scalability. 

To this end, we propose SimC3D, a simple but effective 3D contrastive learning framework, for the first time, pretraining 3D backbones from pure RGB image data. To be specific, SimC3D constructs synthetic point clouds through depth estimation models like MiDaS~\cite{midas}. Although the generated point clouds may be imprecise, suitable data processing is carefully designed, including scale matching, view mixup, and strong 3D augmentations.

We also enhance the pretraining method by studying the learning objective. By rethinking previous multi-modal methods~\cite{invar3D,crosspoint},  our study first reveals 
that the key factor for the multi-modal 3D pretraining is encoding the point cloud's locality, and such approach lack the encoding of the global position awareness.
By using a 2D positional embedding map~\cite{2dpos} as the learning objective, SimC3D further simplifies the pipeline and enhances performance, as shown in Figure~\ref{fig:teaser}(b). 

Remarkably, the proposed SimC3D significantly improves various downstream tasks, e.g., 3D object detection~\cite{votenet,imvotenet} and 3D segmentation~\cite{qi2017pointnet++,3dunet}. Compared with previous works, SimC3D performs comparably or better than state-of-the-art methods~\cite{ponder,invar3D, msc}, at lower cost and without needing real point cloud data. 


%% file: sec/2_related.tex
\section{Related Works}
\subsection{Self-Supervised Visual Pretraining}
In recent years, self-supervised pre-training has emerged as a crucial field for advancing vision tasks, particularly in the domain of 2D images~\cite{simCLR,simsiam,mae}. These pre-training methods can be broadly categorized into two types: discriminative-based~\cite{simCLR} and generative-based~\cite{mae}. Discriminative pre-training designs proxy tasks to guide representation learning, with contrastive learning as a typical example, which employs instance discrimination to establish invariance between multiple augmented views. While generative-based methods focus on reconstructing the original image. These 2D pre-training techniques have yielded impressive improvements in downstream tasks such as object detection and semantic segmentation. As a consequence, many pre-training methods~\cite{pointcontrast,depthcontrast} for 3D models have been inspired by and successfully applied to 3D downstream tasks.

\subsection{Single-Modal 3D Pretraining}
Single-modal methods mostly utilize multi-view information for learning. For discriminative methods, PointConstrast~\cite{pointcontrast} is the first to use multi-view correspondence to pre-train the 3D backbone. They perform different data augmentations on two adjacent views, where the point pairs at the same position in the original space are regarded as positive pairs, and the ones at different positions are regarded as negative samples. Then a standard contrastive loss is used for training. Inspired by this work, several works design different strategies to improve contrastive learning results. CSC~\cite{c_scene}  designs a new loss function to focus on spatial contexts. 4dContrast~\cite{4dcontrast} utilizes 4D signal from moving object sequences and introduces losses among different levels. Other works~\cite{strl,randomroom} try to work on the different formats of data with stronger correspondence.

Different from discriminative training, other methods resort to generative learning, \eg completing the point cloud~\cite{mae3d,mae3dobj,iae,trans3d}. A typical recent work IAE~\cite{iae} uses an implicit decoder to reconstruct the original point cloud from the coarse feature. Rather than using a partial point cloud, they crop the data from the complete point cloud and apply mask augmentation on it.  Despite achieving excellent results on downstream tasks, these methods either need complete point cloud or multi-view correspondence. Thus, it is challenging for them to further scale up the experiments and transfer to different scenarios. DepthContrast~\cite{depthcontrast} employs instance-level contrastive learning to address this problem but requires a much larger batch size (\ie, 4096) and more data compared with others. 

\subsection{Multi-modal 3D Pretraining}
Multi-modal 3D pretraining~\cite{ponder,invar3D,crosspoint,im2point,zhang2022learning,qian2022pix4point} methods are becoming more popular in recent years. These methods try to leverage other modalities to learn representative features, \eg from depth maps and RGB images. Typically, Cross Point~\cite{crosspoint} lets the 3D backbone learn from RGB representations, Invar3D~\cite{invar3D} explores the difference of using pixel-level contrastive learning with different modalities. They empirically show that learning with a depth map gives the best performance. This 3D to 2D learning indeed works better, but it remains unclear why a depth map is better than an RGB image. In addition, the different 2D backbones could also influence the results, thus requiring further investigation. Different from the above methods, there are also some works simultaneously utilizing multi-modal and multi-view information. Ponder~\cite{ponder} follows the idea of IAE but replaces the point cloud reconstruction with depth and RGB image rendering. In this way, they can also leverage the color information during pretraining but have more requirements on data to achieve the best performance.
In comparison, our proposed method is much simpler than existing ones, as it works on single-view data and is easy to combine with other modalities. Specifically, we rethink the key factor of multi-modal 3D contrastive learning and propose a better objective.

%% file: sec/3_method.tex

\input{tab/result}
\section{Rethinking Multi-Modal 3D Contrastive Learning}
\subsection{Unified Multi-Modal Framework}
\label{sec:intro_contrast}
Contrastive 3D pre-training usually adopts two branches. As shown in Figure~\ref{fig:teaser}, typically, the left branch is called the online branch, which contains the 3D backbone to be pre-trained. The target branch is on the right to provide learning objectives.
\
In the online branch, the input point cloud  $\mathcal{X}^{N\times 3}$ is firstly transformed by various 3D augmentations. Then, it is fed into the 3D backbone $f$, where the dimension $3$ means the 3D coordinates. A general 3D backbone like PointNet++~\cite{qi2017pointnetplusplus} extracts the feature of point cloud $\mathcal{X}$. Then, the feature will pass an MLP head $h$ to be projected to invariance space, which is proved to be beneficial by many works~\cite{simCLR,simsiam}. The final representation $q^{N\times C}$ of the point cloud can be written as $q=h(f(\mathcal{X}))$. Then, a traditional contrastive loss like InfoNCE~\cite{infonce} will be applied for pretraining.

In multi-modal methods, the target branch takes inputs like RGB images or Depth images, and a 2D feature map is acquired with the 2D backbone. As the correspondence can be guaranteed by the 3D-2D projection, the corresponding features for each point are sampled from the 2D feature map, thus also resulting in a point-level target feature $k^{N\times C}$.

\subsection{Analysis on 2D Target}
\label{sec:2D_study}
We use the unified framework described in Sec \ref{sec:intro_contrast} to explore the key factor of the target 2D branch. Specifically, we study the difference when choosing various 2D backbones, choosing depth image as the 2D input since it is verified to be a more powerful modality~\cite{invar3D}. Here, our metric is taken as the $\text{AP}_{25}$ results on SUN RGB-D datasets. We start with verifying the performance of 3 pre-trained models: ViT, ResNet50, and ResNet18. Then, we stopped freezing the weights and tested the performance. At last, we directly use a randomly initialized ResNet18 as the backbone. The results are shown in Table~\ref{tab:reduc_models}. 

Previously, an image encoder was considered necessary to extract semantic information to assist with 3D downstream tasks. However, our experiments show that a deeper 2D encoder doesn't guarantee better performance, like ViT does not outperform resnet18 notably (Table~\ref{tab:reduc_models}). To investigate this,  we design a step-by-step reduction procedure to analyze the main factors, shown in Table~\ref{tab:reduc_layers}. We first remove two blocks from ResNet and then remove one. In the last step, we remove all parts from ResNet and only use the output layer to map the input to a higher dimension. Counterintuitively, all these 2D target models result in similar performance, which indicates a potential learning bottleneck.

\subsection{Locality Learning is the Key Objective}


To further study the reason for such a counterintuitive phenomenon, we study the simplest 2D encoder in Table~\ref{tab:reduc_layers}, which consists of a single convolutional layer that projects the image into high-dimensional feature space. This is our most simplified model from deep 2D networks and can only encode the local information. Specifically, we use a large kernel resulting in a $7\times 7$ feature map to be consistent with deep CNN's feature map. Detailed analysis in 3 aspects are as follows:
\begin{itemize}[leftmargin=6mm]
\item\textbf{Semantic information is not crucial:} As shown in Table~\ref{tab:re_onelayer} (c) and (e), replacing 2D branch with distilling 2D features to 3D will cause much lower performance. 
\item\textbf{ 2D backbones share a similar objective:} We visualize the target 2D point features by PCA. In Figure~\ref{fig:pca_reduce}: (1) the learned features of different 2d encoders for the same input points are similar, suggesting a similar learning objective; (2) The ResNet101 feature is similar with different input images, suggesting the learning objective is not strongly related to the semantic content of the input image.

\item\textbf{ Locality is the key for the equivalence:} We decrease the kernel size from 38 to 3 in the single convolutional layer setting. In Table ~\ref{tab:re_onelayer}(a) and (b), without a large kernel to encode the locality, the performance will drop drastically.
\end{itemize}
This empirical analysis shows that locality is an important factor in 3D pretraining and inspired us to encode global awareness further to improve performance. 

Notably, with different 2D inputs, the locality can be divided into RGB locality and depth locality. As depth gives cleaner information about 3D location, it tends to give better performance (Table~\ref{tab:targets}). However, such exploration leads to the fact that there is redundancy in the 2D branch, and the global position awareness is missing. We propose a better and simpler position learning strategy in Sec.~\ref{sec:method_position}.
 \input{fig/pipeline}
\section{SimC3D Methodology}
\subsection{\ours Pipeline Overview}
As shown in Figure~\ref{fig:framework}, the input of our pipeline is an RGB image. We first generate a synthesized point cloud $\mathcal{X}^{N\times 3}$ from it using our RGB-based dataset construction. Then, we use different ways to build the online branch and target branch. The online branch follows the standard 3D forward process, being augmented, passing through the 3D backbone, and projected to invariance feature space through an MLP head. For the target branch, we build a lightweight framework using 2D position as the learning objective. We directly sample fixed 2D positional encoding as a feature and project it through the MLP head. The point and its corresponding 2D position feature are considered positive pairs.

We follow the standard contrastive learning, using InfoNCE loss\cite{infonce} as the main loss function, which can be represented as Eqn.~\ref{eq:contrastive_loss}.
\begin{equation}
\label{eq:contrastive_loss}
	L_{\text{InfoNCE}} = \mathop{\mathbb{E}}_{b}\bigg[\mathop{\mathbb{E}}_{i}\bigg(-\log\frac{\exp{(\mathrm{cos}(q_b^i, k_b^i)/\tau)}}{\sum_{t=1}^N\exp{(\mathrm{cos}(q_b^i, k_b^t)/\tau)}}\bigg)\bigg]
\end{equation}
Where $(q_b^i,k_b^i)$ are corresponding point pairs from the online branch and target branch in the same scene, and $b$ denotes different scenes. $\tau$ is the temperature hyper-parameter.

In the meanwhile, as some 3D backbones~\cite{me} take points' color as additional input, we adopt the color reconstruction loss~\cite{msc} as an additional learning objective for these backbones to fully leverage RGB inputs. This loss is optionally used based on the input requirements of the 3D backbone.


\subsection{RGB-based Dataset Construction}
In addressing the significant reliance on high-precision 3D point clouds, our proposed methodology leverages the construction of point clouds from RGB images. Specifically, for each image, we utilize MiDaS~\cite{midas} to extract its inverse depth map. Subsequently, employing standardized camera intrinsics and depth scale, we project these into point clouds within the camera coordinate system. Due to the unpredictability of the camera extrinsic, we apply a simple rotation of the coordinate axes to transform them into world coordinates. The whole process can be represented as:
\begin{align}
    P_{\text{world}} = R \cdot (K^{-1} \cdot s \cdot D_{\text{MiDaS}}(\text{image}))
\end{align}
Where $D_{\text{MiDaS}}(\text{image})$ represents the inverse depth map extracted using MiDaS, \(K^{-1}\) denotes the inverse of the assumed camera intrinsics, \(s\) is the depth scale, and \(R\) is the rotation matrix for transforming the camera coordinates to world coordinates. 

 Our approach does not entail image-specific adjustments, which ensures the broad applicability of our method to RGB images, circumventing the constraints imposed by the availability of high-precision 3D data. However, the projected point clouds may follow a specific distribution, and directly using them may hurt the performance. 
 
 To ensure the data has enough variety, we use several strategies: (1) \textit{Scale matching:} We adopt the camera parameters from one sample of ScanNet~\cite{scannet}. We also ensure the depth scale is within the common range of the 3D dataset; (2)\textit{View mixup:} Usually, monocular point clouds only represent part of the scene. To represent more complex layouts, we randomly mixup two point clouds with a certain probability. \textit{Strong 3D augmentations:} Variety is one of the most important factor for 3D pretraining. Following previous works~\cite{pointcontrast, msc}, we adopt strong augmentations like random-scale, random-rotate, random-sample, random-crop, and random drop.


\subsection{2D Position Learning Objective}
\label{sec:method_position}
To fully leverage locality learning and make the point feature aware of its global context, we propose a new 2D positional learning objective to replace the target modality. In experiments, it shows better stability and performance without the requirements for 2D input modalities on the target branch. 
As shown in Figure~\ref{fig:framework}, for the 2D modality, we choose to directly use a constant 2D Positional Encoding\cite{2dpos} as the coarse target. Mathematically, positional encoding can be written as:
\begin{equation}
\begin{aligned}
& PE(x,y,2i) = sin(x/10000^{4i/d_{model}}), \\
& PE(x,y,2i+1) = cos(x/10000^{4i/d_{model}}), \\
& PE(x,y,2j+d_{model}/2) = sin(y/10000^{4j/d_{model}}), \\
& PE(x,y,2j+1+d_{model}/2) = cos(y/10000^{4j/d_{model}}),    
\end{aligned}
\end{equation}
where $d_{model}$ is the output dimensions, $x$,$y$ denotes the position and $i$ is the dimension. Our position learning objective significantly improves performance in various downstream tasks.

\noindent\textbf{Understanding from contrastive learning perspective} 
Our 2D position learning has two roles in contrastive learning. Firstly, it serves as a stabilized contrastive learning target. As every point has a fixed 2D position from its original image, their feature is forced to be similar after different augmentation. In addition, different points are encouraged to have different features. The benefit of such a design is to avoid overfitting on correspondence-finding tasks but concentrate on learning discriminative features and learning local geometry. 

The second role of our position learning is to learn the sense of relative position. As the position information is explicitly encoded in the learning target, the network is trained to extract such position information. In addition, As nearby points are more likely to belong to the same class, our positional encoding forces the nearby points to be similar and faraway points to be dissimilar in the feature space. 

%% file: tab/result.tex
\begin{figure}[t] 
    \centering
    \begin{minipage}{0.50\textwidth}
    {
    \centering
    \setlength{\tabcolsep}{1mm}{
    \begin{tabular}{c|cc|c}
    \toprule
    
    \multirow{2}{*}{2D model} & \multirow{2}{*}{trainable} & \multirow{2}{*}{pretrained} & \multicolumn{1}{c}{SUN RGB-D}     \\ \cline{4-4} 
                              &                            &                             & AP25 \\ \hline
                    ViT     &\xmark                      &\cmark       &60.1\\ 
                    ResNet50       &  \xmark                        &    \cmark                            & 60.5     \\ 
                   ResNet18       &  \xmark                        &    \cmark                            & 60.1     \\ 
                  ResNet18       & \cmark                  &   \xmark                               &  60.4    \\ 
                   ResNet18       &  \xmark                        &    \xmark                           & 60.1     \\ 
    \bottomrule
    
    \end{tabular}
    \captionof{table}{\textbf{Analysis of different backbones}: There's no significant difference between different backbones, even for the random initialized and frozen ResNet~\cite{resnet}.}
    \vspace{-5pt}
    \label{tab:reduc_models}
    }
    }
    \end{minipage}
    \hspace{3mm}
    \begin{minipage}{0.45\textwidth}
    \centering
    {
    \setlength{\tabcolsep}{1mm}{
    \begin{tabular}{l|cc|c}
    \toprule
    \multirow{2}{*}{2D model} & \multicolumn{1}{c}{SUN RGB-D}     \\ \cline{2-2} 
                                                 & AP25 \\ \hline
                4 block                                 & 60.4     \\ 
               2 block + output layer                                & 60.4     \\ 
              1 block + output layer                                  &  60.1    \\ 
                 output layer                            & 60.3     \\ 
    \bottomrule
    
    \end{tabular}
    \captionof{table}{\textbf{ResNet18 Reduction}: We gradually reduce the ResNet18 architecture to one simple convolutional layer. The performance is not lowered.}
    
    \label{tab:reduc_layers}
    }}
    \end{minipage}
    \vspace{4mm}
    
    \begin{minipage}{\columnwidth}
        \centering
        \setlength{\tabcolsep}{1mm}{
        {
        \begin{tabular}{ll|c}
        \hline
        &Method & mIoU \\ \hline

        (a)&SimC3D (Color, default Single layer kernal\_size = 38)        & 74.7         \\
        (b)&Single layer, kernal\_size = 3         & 73.0  \\ 
        (c)&Replace Single layer with ResNet101 & 74.5\\ 
        (d)&Replace Single layer with ViT & 74.8\\
        (e)&w/ ResNet101 (pretrained, freezed) distill feature to 3D & 73.5\\
        \hline

        \end{tabular}
        }
        }

        \captionof{table}{\textbf{Semantic Segmentation on ScanNet.}}
        \vspace{-3mm}
        \label{tab:re_onelayer}
    \end{minipage} 

    \begin{minipage}{\columnwidth}
        \centering
        \vspace{-16mm}
        \includegraphics[width=.7\linewidth]{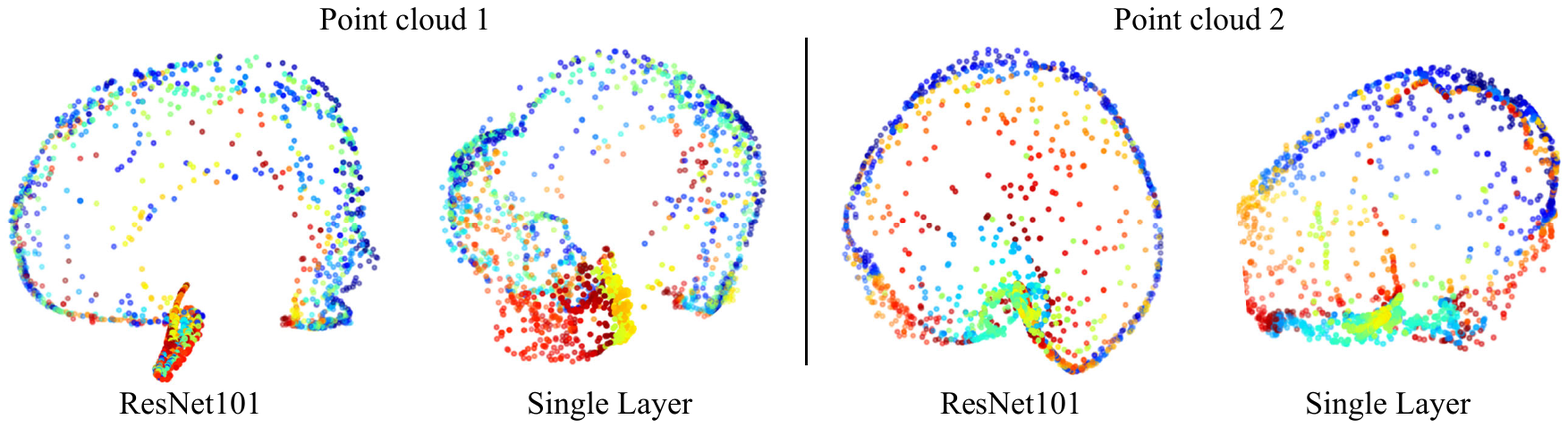}
        \vspace{-23mm}
        \captionof{figure}{\textbf{PCA analysis on points' 2D target feature.}}
        \label{fig:pca_reduce}
    \end{minipage}
\vspace{-2mm}
\end{figure}

%% file: fig/pipeline.tex
\begin{figure*}[t]
\begin{center}
\includegraphics[width=1\textwidth, trim=73mm 30mm 73mm 30mm, clip]{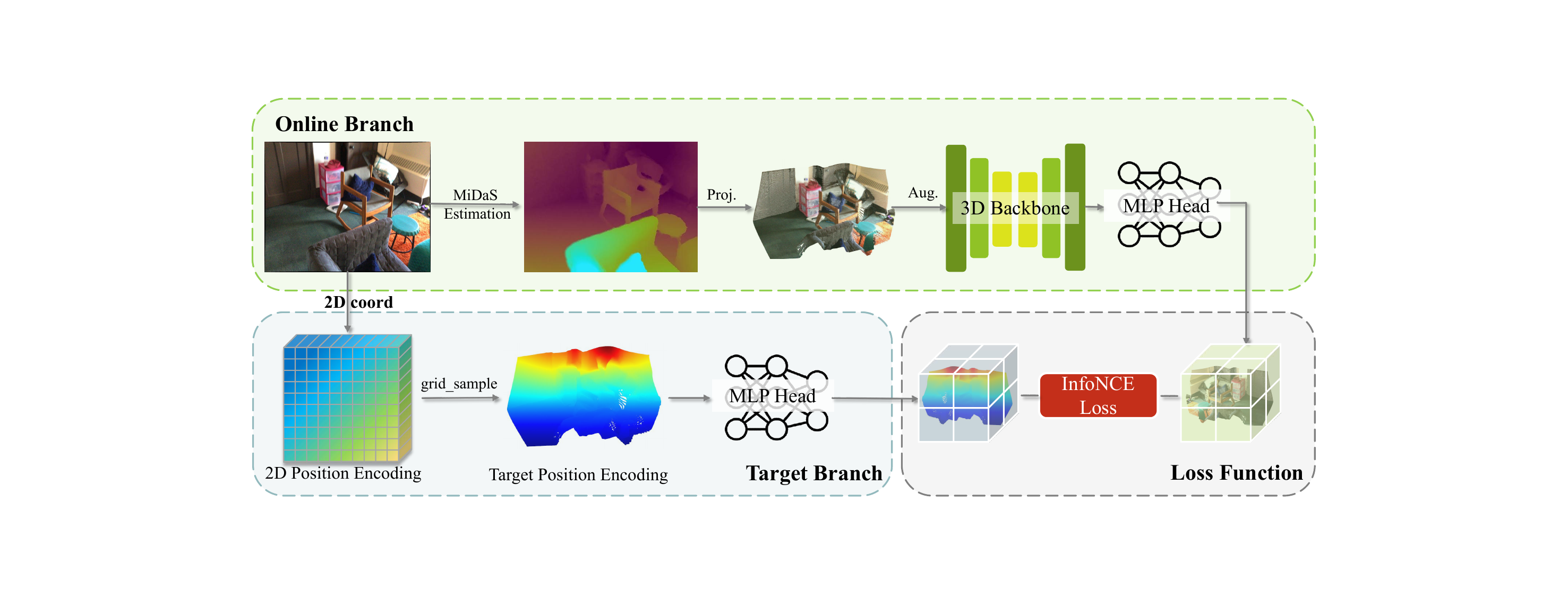}
\end{center} 
\caption{\textbf{SimC3D Overview}: Given an RGB image, we first use MiDaS~\cite{midas} to extract the inverse depth map, and then project it to a point cloud with fixed camera calibration parameters. Then, we adopt a suitable point cloud processing and extract the online branch feature. For the target branch,  we directly apply the 2D positional encoding~\cite{2dpos} in the 2D branch. Each point's target is directly sampled from the feature map by their 2D coordinates. Finally, a contrastive infoNCE loss and a RGB reconstruction loss is used for pre-training}
\vspace{-10pt}
\label{fig:framework}
\end{figure*}

%% file: sec/4_experiment.tex
\section{Experiments}
\input{tab/segment}

\input{tab/instance}
\subsection{Pre-training Setting}
\input{tab/det}
\noindent\textbf{Dataset} 
Our pre-training is based on RGB images from ScanNet Dataset~\cite{scannet}. It is a widely used indoor dataset with more than 1,500 scenes, containing 2.5 million frames. In our pre-training, we sample frames at intervals of 25, resulting in about 100k frames.  We also conducted a scaling experiment on the semantic segmentation task. We use COCO~\cite{coco}2014 training set as an additional data source, containing 83k images.

\noindent\textbf{Augmentations}
Following previous works, We use random sampling, rotation, translation, and cropping for augmentation. For points' color, we use color jittering as the augmentation. In particular, for the semantic segmentation and instance segmentation task, we adopt the random masking augmentation strategies used by MSC~\cite{msc}.

\subsection{Downstream Task Performance}
In this section, we evaluate our SimC3D method for semantic segmentation, instance segmentation and object detection tasks. The evaluation involves Voxel-based SparseUNet~\cite{me} and point-based PointNet++~\cite{me}.

\noindent\textbf{Semantic segmentation }
In Table~\ref{tab:3d_segment}, we evaluate the semantic segmentation performance on both the ScanNet~\cite{scannet} and ScanNet200 benchmarks. Specifically, we utilize SparseUNet as our backbone and compare our results with previously reported performances. Notably, by using only ScanNet as the pretraining dataset, we are able to surpass the state-of-the-art performance of MSC's~\cite{msc}, which utilizes a scaled dataset combining ScanNet and ArkitScenes. Furthermore, we explore the scalability of our method by incorporating additional data from COCO 2014~\cite{coco}. Results show that our method achieves state-of-the-art performance on both benchmarks when supplemented with extra data.

\noindent\textbf{Instance segmentation }
In Table~\ref{tab:instance}, we present our performance in instance segmentation. In this experiment, we also use SparseUnet as the backbone, driven by PointGroup~\cite{pointgroup}. Compared to previous methods, SimC3D demonstrates exceptional performance in instance segmentation. Notably, when compared with the state-of-the-art (SOTA) MSC method, our approach achieves higher scores using only ScanNet data. Specifically, on ScanNet200, our mAP@25 and overall mAP significantly surpasses the SOTA. These experiments demonstrate the full potential of our method in instance segmentation, particularly across a broader range of categories.
\input{tab/sup_efficient}

\noindent\textbf{{Object detection}}
\label{l:sota}
We report the object detection performance in Table~\ref{tab:3d_detect} with PointNet++ Backbone and in Table~\ref{tab:vox_det} with SparseUnet backbone. We conduct the comparison on ScanNet and Sun RGB-D~\cite{sunrgbd} dataset. Following previous works~\cite{pointcontrast,c_scene}, we use average precision with 3D detection IoU threshold 0.25 and threshold 0.5 as the evaluation metrics.  For the PointNet++ backbone, our methods achieve competitive performance compared with previous methods. On the ScanNet dataset, we achieve comparable results with SOTA Ponder~\cite{ponder}, which uses a neural rending way for training and requires multi-view information. On the SUN RGB-D dataset, Our methods achieve comparable results on $\text{AP}_{50}$ metric and get SoTA results on $\text{AP}_{25}$. For the SparseUNet backbone, our method consistently achieves better performance.



\subsection{Data Efficiency}
We evaluate data efficiency on the standard ScanNet~\cite{scannet} benchmark. It includes limited annotations (LA) and limited reconstructions (LR). In LA, only a small subset of points have labels, and in LR, only a small number of scenes can be used during training. We finetune our model, which is pretrained on ScanNet~\cite{scannet}. The results in Table~\ref{tab:seg_eff} show that our method outperforms previous methods~\cite{c_scene, vibus,msc} a lot on the data efficiency benchmark. The improvement is more significant when there is more data or annotations.

\subsection{Ablations}
\noindent\textbf{Major component ablation}
Our data construction simplification and position learning objective can improve performance with lower data requirements. To verify this, we compare the results by replacing synthesized data with real data. We also changed the position learning objective to the original 3D correspondence matching objective. As shown in Table~\ref{tab:vs}, synthesized data from RGB images won't decrease the performance a lot. However, it has worse performance than real data since, probably overfitting to the 3D correspondence objective. With our 2D position learning objective, the pretraining can fully leverage the synthesized data and achieve better performance.

\input{tab/exp_abla_pos}

\noindent\textbf{Ablation of different target format}
 Because our method is compatible with different 2D modalities at the same time, it is necessary to explore different modalities.  For the depth input, we use plotted heat maps instead of raw data. Therefore, the input dimensions of depth maps and RGB images are the same. On the parameters of the convolutional layer, we set its kernel size to 38 and its stride to 32, resulting in a $7\times 7$ feature map. We evaluate the performance of SUN RGB-D object detection tasks.

The result is shown in Table~\ref{tab:targets}. Similar to the finding of Invar3D~\cite{invar3D}, we find that in 2D modalities, the depth image is more effective than the RGB image. Meanwhile, our depth-based method performs comparably to Invar3D's results, which proves that our simplification of the 2D encoder does not affect performance.


\noindent\textbf{Ablation of position objective}
\label{sec:pos_ablation}
As we are using a shared positional encoding, we find that our contrastive method can also be transformed into a regression-based method. Consider using a $7\times 7$ target; We can let the 3d branch directly predict where each point falls, that is, 49 categories of classification. In the implementation, we directly add one classification layer after the 3D branch's projection head and use Cross Entropy loss. Similar to previous experiments, we measure the detection performance on SUN RGB-D.

From the results in Table~\ref{tab:pos_det}, contrastive learning works better in this case. One possible explanation is that the key feature is sampled bilinearly. Thus, the correspondence learned is better compared with discrete classification.

\noindent\textbf{Supplementary}
We offer additional ablation study, analysis, limitations and social impact of our work in the supplementary material.



%% file: tab/segment.tex
\begin{table}[t]
\centering
{\small

\setlength{\tabcolsep}{1mm}{
\begin{tabular}{l|cc|c|c}
\toprule
Method         & source data type     & pretrain dataset      & ScanNet mIoU & ScanNet200 mIoU \\ \hline
SparseUNet~\cite{me}     & -                    & -                     & 72.2         & 25              \\
+ PC~\cite{pointcontrast}           & Raw Point Cloud      & ScanNet               & 74.1         & 26.2            \\
+ CSC~\cite{c_scene}          & Raw Point Cloud      & ScanNet               & 73.8         & 26.4            \\
+ MSC~\cite{msc}          & Complete Point Cloud & ScanNet               & 75.0         & -               \\
+ GC~\cite{gc}          & Complete Point Cloud & ScanNet               & 75.7         & 30.0               \\
+ SimC3D(Ours) & RGB Images           & ScanNet               & \textbf{76.2}\increase{4.0}         & \textbf{30.2}\increase{5.2}            \\ \hline
+ MSC         & Complete Point Cloud & ScanNet + ArkitScenes~\cite{arkit} & 75.5         & 28.8            \\
+ PPT (Unsup.)~\cite{ppt} & Complete Point Cloud & ScanNet + ArkitScenes~\cite{arkit} & 75.8         & 30.4            \\
+ SimC3D (Ours) & RGB Images           & ScanNet + COCO2014~\cite{coco}       & \textbf{76.6}\increase{4.2}         & \textbf{30.7}\increase{5.7}         \\ \bottomrule
\end{tabular}
}}

\caption{\textbf{Semantic Segmentation.} We measure the mIoU score with SparseUNet backbone on ScanNet and ScanNet200 dataset. Compared with previous methods, our ScanNet pretrained model can even beat the MSC method which is pretrained on more scenes. By incorporating CoCo2014 data, we are able to achieve SOTA.}
\vspace{-5mm}
\label{tab:3d_segment}
\end{table}

%% file: tab/instance.tex
\begin{table}[t]
\centering
{\small
\setlength{\tabcolsep}{0.8mm}{
\begin{tabular}{l|c|ccc|ccc}
\toprule

\multirow{2}{*}{Method} & \multirow{2}{*}{pretrain dataset} & \multicolumn{3}{c}{ScanNet}                   & \multicolumn{3}{c}{ScanNet200} \\ \cline{3-8} 
                        &                                   & mAP@25        & mAP@50        & mAP           & mAP@25    & mAP@50    & mAP    \\\hline 
PointGroup~\cite{pointgroup}              & -                                 & 72.8          & 56.9          & 36            & 32.2      & 24.5      & 15.8   \\
+ PC~\cite{pointcontrast}                    & ScanNet                           & -             & 58.0          & -             & -         & 24.9      & -      \\
+ CSC~\cite{c_scene}                  & ScanNet                           & 73.8          & 59.4          & -             & -         & 25.2      & -      \\
+ LGround~\cite{lground}              & ScanNet                           & -             & -             & -             & -         & 26.1      & -      \\
+ MSC~\cite{msc}                  & ScanNet + ArkitScenes             & 74.7          & 59.6          & 39.3          & 34.3      & 26.8      & 17.3   \\
+ GC~\cite{gc}          &  ScanNet               & -         & \textbf{62.3}          & -          & -      & 27.5      & -   \\
+ SimC3D (Ours)          & ScanNet                           & \textbf{76.1}\increase{3.3} & 61.2\increase{4.3} & \textbf{39.9}\increase{3.9} &   \textbf{37.2}\increase{5.0}        &  \textbf{30.6}\increase{6.1}        & \textbf{21.5}\increase{5.7}     \\ \bottomrule
\end{tabular}
}}
\caption{\textbf{Instance Segmentation.} We measure the mAP metric with pretrained SparseUNet backbone, driven by PointGroup~\cite{pointgroup}. Compared with previous methods, Our SimC3D achieves significantly better performance on ScanNet and ScanNet200 validation sets. While only pretrained on ScanNet, SimC3D can beat MSC which is trained on a larger scale dataset. }
\vspace{-3mm}

\label{tab:instance}
\end{table}

%% file: tab/det.tex
\begin{table}[t]
\centering
{\small
\vspace{-5mm}
\setlength{\tabcolsep}{1.8mm}{
\begin{tabular}{l|ccc|cc|cc}
\toprule
\multirow{2}{*}{Method}             & \multirow{2}{*}{Type} & \multirow{2}{*}{Source Data} & \multirow{2}{*}{Epochs} & \multicolumn{2}{c}{ScanNet}                 & \multicolumn{2}{c}{SUN RGB-D}         \\ \cline{5-8} 
                                    &                                    &                                    &                                      & $\text{AP}_{50}$     & $\text{AP}_{25}$     & $\text{AP}_{50}$ & $\text{AP}_{25}$ \\ \hline
VoteNet\cite{votenet}                        & -                                  & -                                  & -                                    & 33.5                 & 58.6                 & 32.9             & 57.7             \\
+ STRL\cite{strl}                                & Contrast                           & Depth                              & 100                                  & 38.4                 & 59.5                 & 35.0             & 58.2             \\
+ RandomRooms\cite{randomroom}                         & Contrast                           & Synthesis                          & 300                                  & 36.2                 & 61.3                 & 35.4             & 59.2             \\
+ PC\cite{pointcontrast}                       & Contrast                           & 3D Model                           &     -                                 & 38.0                 & 59.2                 & 34.8             & 57.5             \\
+ PC-FractalDB\cite{frdb}                        & Contrast                           & Synthesis                          &     -                                 & 38.3                 & 61.9                 & 33.9             & 59.4             \\
+ DepthContrast\cite{depthcontrast}                       & Contrast                           & Depth                              & 1000                                 & 39.1                 & 62.1                 & 35.4             & 60.4             \\
+ \color{lightgray}{Invar3D*\cite{invar3D}  }                           & \color{lightgray}{Contrast}                           & \color{lightgray}{Depth}                              & \color{lightgray}{120}                                  & \color{lightgray}{41.5}                 & \color{lightgray}{64.2}                 & \color{lightgray}{35.6}             & \color{lightgray}{59.8}             \\
+ IAE\cite{iae}                                 & Generative                           & 3D Model                           & 1000                                 & 39.8                 & 61.5                 & \textbf{36.0}             & 60.4             \\

+ Ponder\cite{ponder}                              & Generative                           & 3D Model                           & 100                                  & \textbf{40.9}                 & \textbf{64.2}                 & \textbf{36.1}             & 60.3             \\ 

+ SimC3D (Ours) & Contrast                           & \textbf{RGB}                              & \textbf{20}                                   & \textbf{40.8}\increase{7.3} & \textbf{64.1}\increase{5.5} & \textbf{36.3}\increase{3.4}             & \textbf{61.2}\increase{3.5}             \\

\bottomrule

\end{tabular}
}
}

\caption{\textbf{Object detection with Pointnet++\cite{qi2017pointnet++}}: We measure the $\text{AP}_{25}$ and $\text{AP}_{50}$ on SUN RGB-D\cite{sunrgbd} dataset and ScanNet\cite{scannet} dataset. VoteNet\cite{votenet} is the baseline method and DepthContrast\cite{depthcontrast} result is adopted from Ponder\cite{ponder}.{\color{lightgray}{Gray Methods}} means that they are using a different setting for VoteNet and have a higher baseline. Our SimC3D reaches comparable results with SoTA on the ScanNet dataset and outperforms other methods on the SUN RGB-D dataset}
\label{tab:3d_detect}
\vspace{-3mm}
\end{table}

%% file: tab/sup_efficient.tex
\begin{table}[tb]
\centering
\begin{minipage}{0.50\textwidth}

{\small
\setlength{\tabcolsep}{1mm}{
\begin{tabular}{c|ccccc}
\toprule
\multirow{2}{*}{Pct.} & \multicolumn{5}{c}{Limited Reconstructions}           \\ \cline{2-6} 
                      & Scratch & CSC & VIBUS & MSC & \textbf{SimC3D} \\ \hline
1\%                   & 26.0    & 28.9 & 28.6  & 29.2 & \textbf{30.7}         \\
5\%                   & 47.8    & 49.8 & 47.4  & 50.7 & \textbf{53.3}         \\
10\%                  & 56.7    & 59.4 & 60.5  & 61.0 & \textbf{64.1}         \\
20\%                  & 62.9    & 64.6 & 64.8  & 64.9 & \textbf{67.3}         \\
100\%                 & 72.2    & 73.8 & -     & 75.3 & \textbf{76.2}     \\
\bottomrule
\end{tabular}
}
}
\label{tab:seg_lr}

\end{minipage}
\begin{minipage}{0.45\textwidth}
{\small
\setlength{\tabcolsep}{1mm}{
\begin{tabular}{c|ccccc}
\toprule
\multirow{2}{*}{Pts.} & \multicolumn{5}{c}{Limited Annotations}               \\ \cline{2-6} 
                      & Scratch & CSC  & VIBUS & MSC  & \textbf{SimC3D} \\ \hline
20                    & 41.9    & 55.5 & 61.0  & 61.2 &  \textbf{61.5}                     \\
50                    & 53.9    & 60.5 & 65.6  & 66.8 & \textbf{68.1}         \\
100                   & 62.2    & 65.9 & 68.9  & 69.7 & \textbf{71.1}         \\
200                   & 65.5    & 68.2 & 69.6  & 70.7 & \textbf{72.3}         \\
Full                  & 72.2    & 73.8 & -     & 75.3 & \textbf{76.2}        \\
\bottomrule
\end{tabular}
}
}
\end{minipage}
\vspace{1mm}
\caption{\textbf{Data efficiency with limited annotations and reconstructions.}We evaluate the SparseUNet~\cite{me} on semantic segmentation tasks following the ScanNet Data Efficient Benchmark. the mIoU metric is used for comparison. Pct. denotes the percentage of scene reconstruction that could be used for training and Pts. denotes the number of annotated points for each scene. }
\label{tab:seg_eff}
\vspace{-6mm}
\end{table}

%% file: tab/exp_abla_pos.tex
\begin{table}[t]
\vspace{3mm}
\centering

\begin{minipage}{0.44\textwidth}
    \centering
\setlength{\tabcolsep}{1mm}{
\begin{tabular}{l|cc}
\toprule
\multirow{2}{*}{Method} & \multicolumn{2}{c}{SUN RGB-D}  \\ \cline{2-3} 
                      &   $\text{AP}_{50}$ & $\text{AP}_{25}$   \\ \hline
From Scratch           &31.7 &55.6    \\ 
PointContrast~\cite{pointcontrast}& 34.8 & 57.5   \\ 
CSC~\cite{c_scene}            & 36.4 & 58.9    \\ 
PC-FractalDB~\cite{frdb}            & 35.9 & 57.1   \\ 
\ours (Ours)                   & \textbf{37.2}\increase{5.5} &\textbf{59.5}\increase{3.9}   \\ 
\bottomrule
\end{tabular}
}
\caption{\textbf{Object Detection with SparseUNet.}  Our method outperforms previous works }

\label{tab:vox_det}
\end{minipage}
\hspace{3mm}
\begin{minipage}{0.5\textwidth}
    \centering
\setlength{\tabcolsep}{1mm}\small{
\begin{tabular}{ccc}
\toprule
data source & objective & ScanNet mIoU \\ \hline
real                         & 3D correspondence             & 75.0                          \\
synthesized                  & 3D correspondence             & 75.2                          \\
real                         & 2D position                & 75.8                          \\
synthesized                  & 2D position                & 76.2                          \\ \bottomrule
\end{tabular}
}
\caption{\textbf{Major component ablation.} Our synthesized data only introduces a minor performance drop for the traditional method. The 2D position objective can better leverage the synthesized data.}
\label{tab:vs}
\end{minipage}
\begin{minipage}{0.50\textwidth}
\centering
{
\setlength{\tabcolsep}{1mm}{

\begin{tabular}{l|cc}
\toprule
\multirow{2}{*}{Method} & \multicolumn{2}{c}{SUN RGB-D}  \\ \cline{2-3} 
                       & AP50 & AP25    \\ \hline
From Scratch            & 32.9 &57.7    \\ 
Invar3D            & 35.6 & 59.8    \\ 
\ours (Color)           & 34.3 & 59.3    \\ 
\ours (Depth)                     & \textbf{36.3}   &60.4 \\ 
\ours (Position)                   & \textbf{36.2} & \textbf{61.2}   \\ 
\bottomrule
\end{tabular}


\caption{\textbf{Ablation of different targets}: SimC3D (Position) perform the best. Our Depth locality version beats the Invar3D~\cite{invar3D} who use a U-Net backbone for depth.}
\label{tab:targets}
}}
\end{minipage}
\hspace{3mm}
\begin{minipage}{0.45\textwidth}
\centering
{
\setlength{\tabcolsep}{1mm}{
\
\begin{tabular}{l|cc}
\toprule
\multirow{2}{*}{Method} & \multicolumn{2}{c}{SUN RGB-D}  \\ \cline{2-3} 
                      &   AP50 & AP25    \\ \hline
From Scratch           &32.9 &57.7    \\ 
Position Classification          & 35.0 & 60.2   \\ 
Position Contrast                  & \textbf{36.2} &\textbf{61.2}   \\ 
\bottomrule
\end{tabular}

\caption{\textbf{Position Learning Ablation.} Compared with the variant of position classification, Position contrast shows better performance}
\label{tab:pos_det}
}
}
\end{minipage}
\vspace{-5mm}
\end{table}

%% file: sec/5_conclusion.tex
\section{Conclusion}
In this paper, for the first time, we show that complex 2D encoders are not necessary for 3D multi-modal pre-training. Based on this, we propose 2D position learning as a more compact and effective pretraining objective. In addition, we also propose a simple but effective way to construct 3D pretraining data without hurting the performance. Compared with previous methods, Our SimC3D can directly act on RGB images with less computation overhead, and with higher or comparable performance than state-of-the-art. As our method has shown great potential for scalability, experiments on larger image sets and generative models could be future directions.


%% file: sec/X_suppl.tex
\appendix
{\Large\centering\bf SimC3D: A Simple Contrastive 3D Pretraining Framework Using Solely RGB Images\\}

\vspace{3mm}
{\large\centering\rmfamily Supplementary Material\\}
\setcounter{section}{0}
\setcounter{figure}{0}
\setcounter{table}{0}
\appendix
\renewcommand{\thefigure}{\Roman{figure}}
\renewcommand{\thetable}{\Roman{table}}
\section{Code Release}
\label{sec:a}
The code and pretrained model will be publicly available on Github upon publication.
\section{Additional Exepriments}
\label{sec:b}
\subsection{Data Efficiency}
We compare the data efficiency on the ScanNet dataset in Figure~\ref{fig:justification}. Our SimC3D shows consistently better results.

\noindent\textbf{Limited scenes}
To test the performance of our method under different amounts of data, we split out \{1\%, 5\%, 10\%, 20\%\} scenes from the official train split on ScanNet~\cite{scannet}. Each split directly takes the corresponding percentage of data according to the scan order. Therefore, the data under each split is the same, and the small split must be included by the large split. We verify the PointNet++~\cite{qi2017pointnet++} model on this metric. As Figure \ref{fig:li_scenes} shows, our method can boost performance when less data is given. Especially when 10\% data is given,  we can improve the performance from 19\% to 35\% on Scannet.

\noindent\textbf{Limited annotations}
We choose the limited points annotations setting from the ScanNet-LA dataset and measure the detection performance. Under this condition, only a few boxes are provided during training. We verify the performance of the PointNet++ model on \{1, 2, 4, 7\} boxes per scene. Detailed results are shown in Figure \ref{fig:li_boxes}. Compared with training from scratch, our pre-trained model can improve performance a lot. The improvements get much higher when more boxes are available.
\input{fig/efficient}
\input{sup_fig/vis_other}
\subsection{Ablation Study}
\noindent\textbf{Different Positional Encoding} In the paper we use a pre-defined 2D positional encoding. While in the application, there are different kinds of positional encoding, \eg, learnable positional embedding, and 1D positional encoding. Here, we compare the difference when changing the type of the embedding in Table ~\ref{tab:pos_ablation}.

\input{tab/dif_pos}
\input{tab/combine}
For pre-defined positional encoding, we compare the 1D encoding and 2D encoding, where the encoding is the same along $x$ coordinates for 1D. We find out that 2D information is necessary to learn a good representation. In addition, when using learnable embedding to replace the fixed positional encoding, there are no noticeable improvements in the performance. Thus, we choose to use a pre-defined positional encoding for simplicity. We also tried to remove the projection head, however, the performance decreases a lot. Thus, though we only need a 2D position learning target, the projection head is essential to encode detailed position information.

    \noindent\textbf{Combination of targets.} A natural idea when adopting different targets is that it's possible to utilize multiple targets and get better performance. We conduct an ablation study on the combination of targets in Table ~\ref{tab:com_ab}. The results show that though different modalities can be utilized through SimC3D with almost no additional computational cost, it's hard to gain improvements from them. When adding depth images or color images, the performance is worse than directly using a positional encoding. One possible reason is that there may be some conflicts for different modalities' training. Thus, a more carefully designed loss needs to be used to benefit from multiple modalities.

\input{sup_fig/sim}
\section{Additional Analysis}
\label{sec:c}
\subsection{Behavior of SimC3D}
Different from traditional contrastive learning where the learning targets are features from the same format, SimC3D has a shared 2D position feature as the learning target. To analyze whether SimC3D has learned discriminative features, we plot the positive and negative similarities for different methods. Specifically, we use the pretrained backbone to calculate the cosine feature similarity between two augmented views from one point cloud. The positive similarity denotes the average cosine positive similarity among all point pairs, and the negative similarity denotes the average cosine similarity between all non-paired points. We conduct the comparison between MSC~\cite{msc} and SimC3D. As shown in Figure~\ref{fig:similarity}, SimC3D can learn a discriminative feature, but not overfit to the correspondence learning task like MSC. 

\input{tab/im_details}
\subsection{Additional visualizations}
To better understand the learned representation for each method, we build a visualization analysis on the extracted features. Specifically, we use two methods: K-means and PCA. K-means is used to show different parts of the point cloud, and PCA can give an overall view of the feature.

As shown in Figure~\ref{fig:vis_other}. Compared with point-level contrastive methods like PointContrast~\cite{pointcontrast} and CSC~\cite{c_scene}, our SimC3D further shows the benefits of position learning. In the second scene's PCA visualization, we can see that PointContrast discriminates the feature too much, making the features of the table to be distinct, but similar to part of the floor. This is because traditional contrastive learning can't leverage the relative position information, leading to overfitting on the correspondence-finding task. Although CSC tries to address this by scene contexts, it still leads to incorrect optimization, e.g. It is highly biased to the height information in both scenes. For generative methods like IAE~\cite{iae}, since it focuses on the reconstruction objective, the model is overfitted to identify the local surfaces. In the second scene, planes in the same direction have similar features. This learned feature is too local, ignoring the object-level information.

While previous methods either overfit on correspondence among global points or local surface information, our simple 2D position learning shows more benefits. As shown in these two scenes, our method can clearly distinguish global information, e.g. the ground. In the meanwhile, different objects or parts show different features, e.g. the cup, table, and background in the second scene.

\section{Implementation Details}
\label{sec:d}
\subsection{Training Setting}
The pretraining and finetuning settings are shown in Table~\ref{tab:setting}. The SparseUnet is finetuned on semantic segmentation, instance segmentation, and object detection tasks. The PointNet++ model is only finetuned on the object detection task. 

We mainly follow MSC~\cite{msc} for the data augmentation strategies, which include color augmentations and coordinates augmentations. The point will be grid sampled by size 0.02 during SparseUnet pretraining. For PoinetNet++, there's no color input and no grid sampling.

For time consumption, training SparseUnet takes 30 hours with 8 A100 GPU and training the PointNet++ model only takes 8 hours with one A100 GPU. 16 workers are used per GPU.


\subsection{Point Cloud Projection}
As we synthesize point clouds from RGB images, we need to estimate several parameters from the RGB images for back projection. Specifically, it includes the depth map, camera intrinsic parameters, and camera extrinsic parameters. For the intrinsic matrix $K$, we choose one camera intrinsic parameter from ScanNet~\cite{scannet} and use it for all the data, which is: 
\begin{align}
    \begin{bmatrix}
574\times(\text{width}/1296) & 0 & 324\\ 
0 & 575\times(\text{height}/968) & 241\\ 
0 & 0 & 1
\end{bmatrix}
\end{align}

Where $(\text{width}, \text{height})$ is the input RGB image's shape.

For the depth map, we use the DPT Large variant from MiDaS~\cite{midas} to extract the inverse depth map $W'$. Then, we use $K_{0,0}\times0.01$ as the scale factor and clip the result to $[0,6]$ to have a similar scale with ScanNet data. The final depth $W$ is calculated as:
\begin{align}
    W=\text{clip}_{[0,6]}(K_{0,0}\times 0.01/W')
\end{align}

Finally, for the extrinsic matrix, we just use an identical matrix. The camera coordinates system is transformed through exchanging the $Y$ and $Z$ axis. Following the approach used in \cite{c_scene, pointcontrast}, we project the point cloud onto the world coordinate system. Let $K^{3\times 3}$ represent the camera intrinsic parameters, and $T^{3\times 4}$ denote the extrinsic parameters. We can split $K^{3\times 4}$ into $[R|t]$, where $R^{3\times 3}$ is the rotation matrix and $t^{3\times 1}$ is the translation matrix.
 Consider $(u, v)$ as the 2D coordinates and $(X, Y, Z)$ as the corresponding 3D points. Assuming this pixel's depth to be $w$, we can establish the following relationship:
    \begin{align}
w\begin{bmatrix}
u\\ 
v\\ 
1
\end{bmatrix}=K(R\begin{bmatrix}
X\\ 
Y\\ 
Z
\end{bmatrix}+t)
    \end{align}

Thus, the 3D point can be calculated as :
\begin{align}
    \begin{bmatrix}
X\\ 
Y\\ 
Z
\end{bmatrix}=R^{-1}(K^{-1}w\begin{bmatrix}
u\\ 
v\\ 
1
\end{bmatrix}-t)
\end{align}

\section{Limitation}
While our work makes a substantial effort to uncover the key factor for 3D pertaining, we find that it still faces limitations. Firstly, our method is mainly for indoor scenes since indoor and outdoor are separately studied in this field. When we construct our data from RGB images, we often obtain dense point cloud, which has a similar distribution to real indoor point clouds. Therefore, our method probably can't be directly applied to outdoor scenes due to their sparseness.

\section{Social Impact}
For the first time \ours propose a method to use RGB image data for pertaining. Such an approach can greatly reduce the social cost of capturing expensive RGB-D data for pertaining and improve the 3D pertaining performance. Thus, it could facilitate complex 3D scene systems and help applications like robots and autonomous driving. Eventually, it could help to build a large foundation model in 3D and make production easier and more efficient. For potential negative impacts, our pretraining data is based on a depth prediction model, some attacks based on these models's behavior could affect the system's performance.

\section{Discussion}
\label{sec:e}
In this paper, we address two unresolved issues in 3D pretraining: the necessity of expensive 3D data and the impact of training across various modalities. Regarding data, there are significant differences between synthetic and real point cloud data. These differences manifest in aspects such as divergent coordinate scales and distortions in the foreground and background. However, we observed that synthesized data can still benefit 3D pretraining. One explanation is that it's essential to learn different local geometry and global layouts for pretraining aimed at dense prediction tasks. Therefore, diversity is more crucial.

In our locality learning, we find that 2D modalities provide local relevance information, indicating which point features are similar or different. Consequently, this objective can be effectively substituted with more explicit 2D positions and improve the performance by its global awareness. Learning 2D positions is a challenging task to some extent, as it involves inferring the shape of a monocular point cloud in its original view without known camera parameters.

Despite the encouraging performance, our method is just the beginning of exploring RGB data in 3D pretraining. Given the numerous benefits of RGB images, such as large amounts and diverse scenarios, it is worth delving deeper into better data synthesis and scaling-up strategies. Specifically, a better projection approach and further augmentations can be applied to enrich the point cloud data. Furthermore, the recently proposed PPT~\cite{ppt} method, which introduces an additional module to benefit large-scale 3D pretraining, can also be attached to our method to further improve scalability.

%% file: fig/efficient.tex
\begin{figure}[h]
\begin{subfigure}{.48\textwidth}
    \centering
    \includegraphics[width=1.0\textwidth]{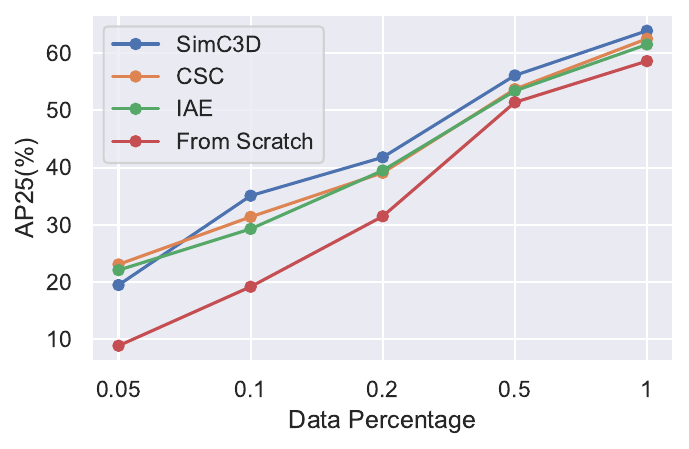}
    \vspace{-1.5em}
    \subcaption{Limited data}
    \label{fig:li_scenes}
\end{subfigure}
\begin{subfigure}{.48\textwidth}
    \centering
    \includegraphics[width=1.0\textwidth]{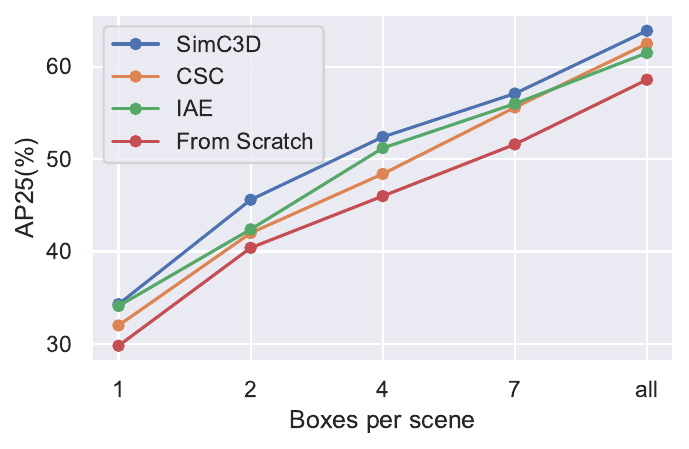}
    \vspace{-1.5em}
    \subcaption{Limited annotation}
    \label{fig:li_boxes}
\end{subfigure}

    \caption{\textbf{Data efficiency comparison.} We conduct comparisons on limited scenes and limited annotations, using Sun RGB-D detection benchmark and PointNet++ as the backbone. Our \ours method shows consistently better results.}
    \label{fig:justification}
    \vspace{-5mm}
\end{figure}

%% file: sup_fig/vis_other.tex
\begin{figure*}[t]
\centering
\includegraphics[width=0.98\linewidth, trim=90mm 64mm 70mm 80mm, clip]{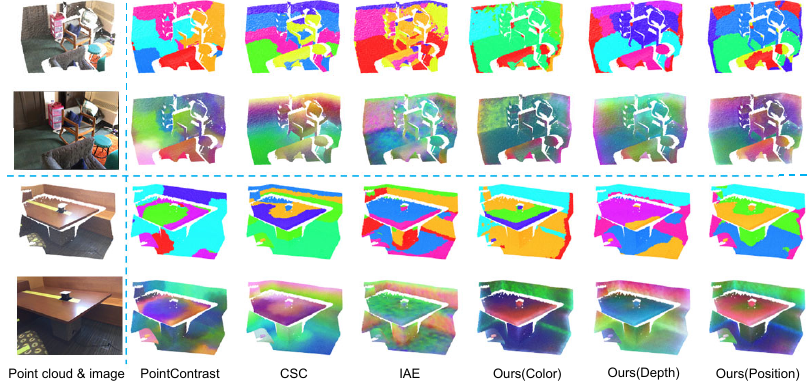}
\vspace{-3mm}
   \caption{\textbf{Additional Visualization Results.} SimC3D achieves better ground awareness and object awareness}
\vspace{-3mm}
\label{fig:vis_other}
\end{figure*}

%% file: tab/dif_pos.tex
\begin{table}[ht]
\centering
\begin{tabular}{c|c}
\toprule
Method & SUN RGB-D  \\ \hline
From Scratch           &55.6    \\ 
1D positional Encoding  & 59.6   \\ 
2D positional Encoding             & 61.2    \\ 
Learnable Embedding            & 60.8   \\ 
Learnable Embedding(no proj head)            & 59.9   \\ 
\bottomrule
\end{tabular}
\caption{\textbf{Positional encoding ablation} with PointNet++~\cite{qi2017pointnet++} on SUN RGB-D detection task. AP25 scores are reported}
\vspace{-5pt}
\label{tab:pos_ablation}
\end{table}

%% file: tab/combine.tex
\begin{table}[ht]
\centering
\begin{tabular}{ccc|c}
\toprule
Position & Depth & Color & SUN RGB-D
\\\hline
            \cmark      &\xmark                      &\xmark       &61.2\\ 
            \cmark       &  \cmark                        &    \xmark                            & 60.0     \\ 
           \cmark       &  \xmark                        &    \cmark                            & 59.9     \\ 
           \cmark       &  \cmark                        &    \cmark                            & 60.8     \\ 

\bottomrule

\end{tabular}
\caption{\textbf{Combination of targets}: utilizing other modalities can't improve the performance for a single modality during pre-training. AP25 scores are reported.}
\vspace{-5mm}
\label{tab:com_ab}
\end{table}

%% file: sup_fig/sim.tex
\begin{figure}[h]
\begin{subfigure}{.48\textwidth}
    \centering
    \includegraphics[width=1.0\textwidth]{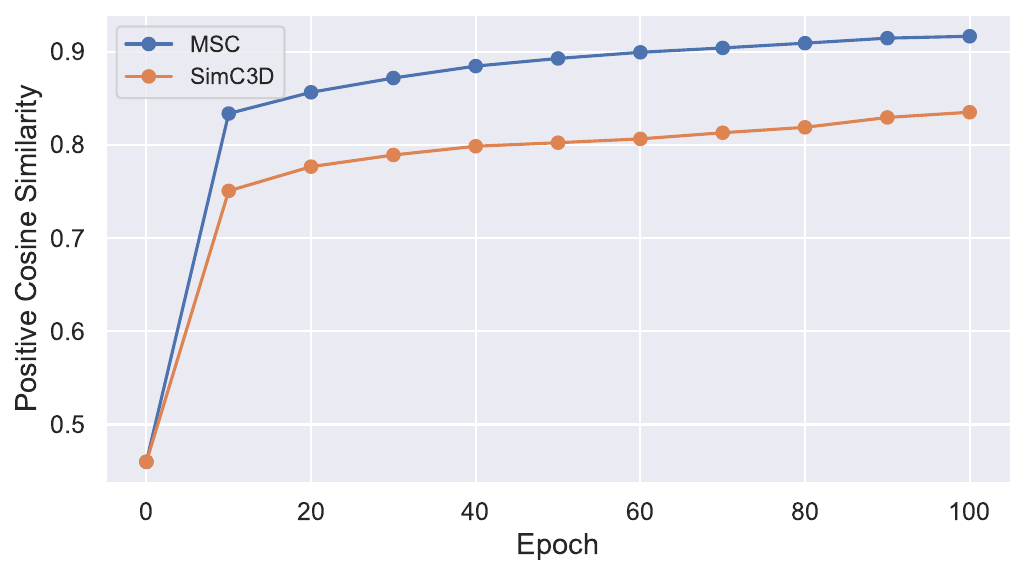}
    \vspace{-1.5em}
    
    \label{fig:pos_sim}
\end{subfigure}
\begin{subfigure}{.48\textwidth}
    \centering
    \includegraphics[width=1.0\textwidth]{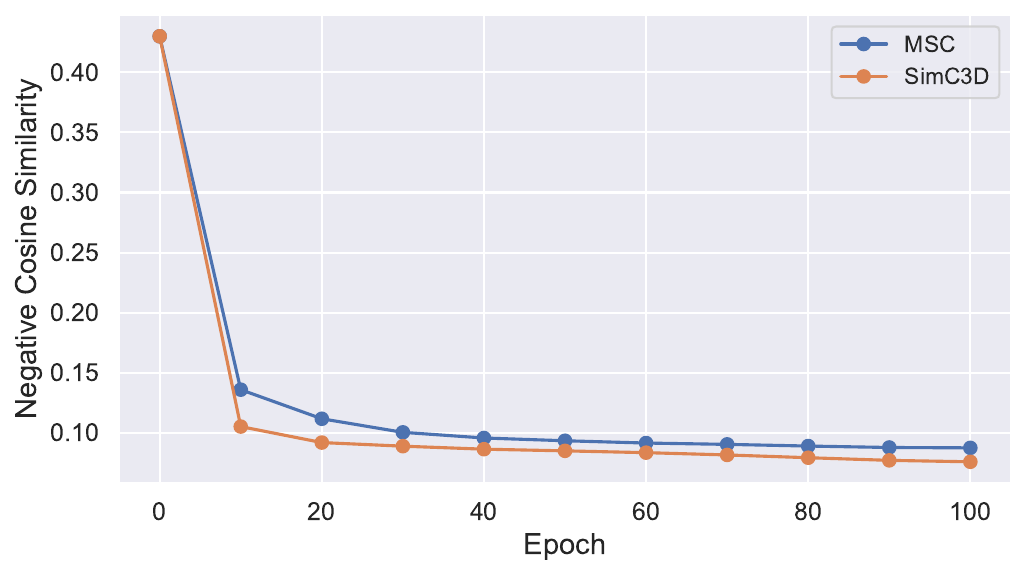}
    \vspace{-1.5em}

    \label{fig:neg_sim}
\end{subfigure}
    \vspace{-0.5em}
    \caption{\textbf{Cosine Similarity Curve.} SimC3D avoids overfitting on the correspondence finding task}
    \label{fig:similarity}
    \vspace{-1mm}
\end{figure}

%% file: tab/im_details.tex
\begin{table*}[t]
\centering
\setlength{\tabcolsep}{0.15mm}{
{\small
\begin{tabular}{cc|cc|cc|cc}
\toprule
\multicolumn{2}{c|}{Pretraining (SP-UNet)} & \multicolumn{2}{c|}{Pretraining (PointNet++)} & \multicolumn{2}{c|}{Finetuning (SP-UNet)} & \multicolumn{2}{c}{Finetuning (PointNet++)} \\ \hline
Config                    & Value             & Config                    & Value             & Config                   & Value             & Config                 & Value              \\
optimizer                 & SGD               & optimizer                 & SGD               & optimizer                & SGD               & optimizer              & Adam               \\
scheduler                 & cosine       & scheduler                 & cosine       & scheduler                & cosine       & scheduler              & step          \\
learning rate             & 0.1               & learning rate             & 0.2               & learning rate            & 0.05              & learning rate          & 1e-3               \\
weight decay              & 1e-4              & weight decay              & 1e-4              & weight decay             & 1e-4              & weight decay           & 0                  \\
momentum        & 0.8               & momentum        & 0.9               & momentum       & 0.9               & -                      &            -        \\
InfoNCE t       & 0.4               & InfoNCE temperature       & 0.07              & -                        &         -          & -                      &       -             \\
batch size                & 32                & batch size                & 64                & batch size               & 48                & batch size             & 8                  \\
warmup epochs             & 1                 & warmup epochs             & 0                 & warmup epochs            & 40                & warmup epochs          & 0                  \\
epochs                    & 100               & epochs                    & 20                & epochs                   & 800               & epochs                 & 180        \\
\bottomrule
\end{tabular}
}
}
\caption{\textbf{Training Settings}}
\label{tab:setting}
\vspace{-5mm }
\end{table*}